\documentclass{article}

\usepackage[final]{nips_2017}

\usepackage[utf8]{inputenc} 
\usepackage[T1]{fontenc}    
\usepackage{hyperref}       
\usepackage{url}            
\usepackage{booktabs}       
\usepackage{amsfonts}       
\usepackage{nicefrac}       
\usepackage{microtype}      
\usepackage{galois}

\usepackage{mathrsfs}
\usepackage{enumerate}
\usepackage{graphicx}
\usepackage{multirow}
\usepackage{amsmath}
\usepackage{epstopdf}
\usepackage{mathrsfs}
\usepackage{amsthm}
\usepackage{amssymb}
\usepackage{appendix}
\newtheorem{theorem}{Theorem}

\title{The Expressive Power of Neural Networks: A View from the Width}

\author{
  Zhou Lu$^{1,3}$\\
  \texttt{1400010739@pku.edu.cn} \\
  \And
  Hongming Pu$^{1}$\\
  \texttt{1400010621@pku.edu.cn} \\
  \And
  Feicheng Wang$^{1,3}$\\
  \texttt{1400010604@pku.edu.cn} \\
  \And
  Zhiqiang Hu$^{2}$\\
  \texttt{huzq@pku.edu.cn} \\
  \And
  Liwei Wang$^{2,3}$\\
  \texttt{wanglw@cis.pku.edu.cn} \\
  \and
  \\
  1, Department of Mathematics, Peking University\\
  2, Key Laboratory of Machine Perception, MOE, School of EECS, Peking University\\
  3, Center for Data Science, Peking University, Beijing Institute of Big Data Research\\
}

\begin{document}

\maketitle

\begin{abstract}
The expressive power of neural networks is important for understanding deep learning. Most existing works consider this problem from the view of the depth of a network. In this paper, we study how width affects the expressiveness of neural networks. Classical results state that \emph{depth-bounded} (e.g. depth-$2$) networks with suitable activation functions are universal approximators. We show a universal approximation theorem for \emph{width-bounded} ReLU networks: width-$(n+4)$ ReLU networks, where $n$ is the input dimension, are universal approximators. Moreover, except for a measure zero set, all functions cannot be approximated by width-$n$ ReLU networks, which exhibits a phase transition. Several recent works demonstrate the benefits of depth by proving the depth-efficiency of neural networks. That is, there are classes of deep networks which cannot be realized by any shallow network whose size is no more than an \emph{exponential} bound. Here we pose the dual question on the width-efficiency of ReLU networks: Are there wide networks that cannot be realized by narrow networks whose size is not substantially larger? We show that there exist classes of wide networks which cannot be realized by any narrow network whose depth is no more than a \emph{polynomial} bound. On the other hand, we demonstrate by extensive experiments that narrow networks whose size exceed the polynomial bound by a constant factor can approximate wide and shallow network with high accuracy. Our results provide more comprehensive evidence that depth may be more effective than width for the expressiveness of ReLU networks.
\end{abstract}

\section{Introduction}

Deep neural networks have achieved state-of-the-art performance in a wide range of tasks such as speech recognition, computer vision, natural language processing, and so on.  Despite their promising results in applications, our theoretical understanding of neural networks remains limited.  The expressive power of neural networks, being one of the vital properties, is crucial on the way towards a more thorough comprehension.

The expressive power describes neural networks' ability to approximate functions. This line of research dates back at least to 1980's. The celebrated universal approximation theorem states that depth-$2$ networks with suitable activation function can approximate any continuous function on a compact domain to any desired accuracy [1][3][6][9]. However, the size of such a neural network can be exponential in the input dimension, which means that the depth-$2$ network has a very large width.

From a learning perspective, having universal approximation is just the first step. One must also consider the efficiency, i.e., the size of the neural network to achieve approximation. Having a small size requires an understanding of the roles of depth and width for the expressive power. Recently, there are a series of works trying to characterize how depth affects the expressiveness of a neural network . [5] show the existence of a $3$-layer network, which cannot be realized by any $2$-layer to more than a constant accuracy if the size is subexponential in the dimension. [2] prove the existence of classes of deep convolutional ReLU networks that cannot be realized by shallow ones if its size is no more than an exponential bound. For any integer $k$, [14] explicitly constructed networks with $O(k^3)$ layers and constant width which cannot be realized by any network with $O(k)$ layers whose size is smaller than $2^k$. This type of results are referred to as depth efficiency of neural networks on the expressive power: a reduction in depth results in exponential sacrifice in width. However, it is worth noting that these are existence results. In fact, as pointed out in [2], proving existence is inevitable; There is always a positive measure of network parameters such that deep nets can't be realized by shallow ones without substantially larger size. Thus we should explore more in addition to proving existence.

Different to most of the previous works which investigate the expressive power in terms of the depth of neural networks, in this paper we study the problem from the view of \emph{width}. We argue that an integration of both views will provide a better understanding of the expressive power of neural networks.

Firstly, we prove a universal approximation theorem for width-bounded ReLU networks. Let $n$ denotes the input dimension, we show that width-$(n+4)$ ReLU networks can approximate any Lebesgue integrable function on $n$-dimensional space with respect to $L^1$ distance. On the other hand, except for a zero measure set, all Lebesgue integrable functions cannot be approximated by width-$n$ ReLU networks, which demonstrate a phase transition. Our result is a dual version of the classical universal approximation theorem for depth-bounded networks.

Next, we explore quantitatively the role of width for the expressive power of neural networks. Similar to the depth efficiency, we raise the following question on the width efficiency:

\emph{Are there wide ReLU networks that cannot be realized by any narrow network whose size is not substantially increased?}

We argue that investigation of the above question is important for an understanding of the roles of depth and width for the expressive power of neural networks. Indeed, if the answer to this question is \emph{yes}, and the size of the narrow networks must be \emph{exponentially} larger, then it is appropriate to say that width has an equal importance as depth for neural networks.

In this paper, we prove that there exists a family of ReLU networks that cannot be approximated by narrower networks whose depth increase is no more than \emph{polynomial}. This polynomial lower bound for width is significantly smaller than the exponential lower bound for depth. However, it does not rule out the possibility of the existence of an exponential lower bound for width efficiency. On the other hand, insights from the previous analysis suggest us to study if there is a polynomial upper bound, i.e., a polynomial increase in depth and size suffices for narrow networks to approximate wide and shallow networks. Theoretically proving a polynomial upper bound seems very difficult, and we formally pose it as an open problem. Nevertheless, we conduct extensive experiments and the results demonstrate that when the depth of the narrow network exceeds the polynomial lower bound by just a constant factor, it can approximate wide shallow networks to a high accuracy. Together, these results provide more comprehensive evidence that depth is more effective for the expressive power of ReLU networks.

Our contributions are summarized as follows:
\begin{itemize}
\item
We prove a Universal Approximation Theorem for Width-Bounded ReLU Networks. We show that any Lebesgue-integrable function $f$ from $\mathbb{R}^n$ to $\mathbb{R}$ can be approximated by a fully-connected width-$(n + 4)$ ReLU network to arbitrary accuracy with respect to $L^1$ distance. In addition, except for a negligible set, all functions $f$ from $\mathbb{R}^n$ to $\mathbb{R}$ cannot be approximated by any ReLU network whose width is no more than $n$.
\item
We show a width efficiency polynomial lower bound. For integer $k$, there exist a class of width-$O(k^2)$ and depth-2 ReLU networks that cannot be approximated by any width-$O(k^{1.5})$ and depth-$k$ networks. On the other hand, experimental results demonstrate that networks with size slightly larger than the lower bound achieves high approximation accuracy.

\end{itemize}

\subsection{Related Work}
Research analyzing the expressive power of neural networks date back to decades ago.  As one of the most classic work, Cybenko [3] proved that a fully-connected sigmoid neural network with one single hidden layer can universally approximate any continuous univariate function on a bounded domain with arbitrarily small error. Barron [1], Hornik et al.[9] ,Funahashi [6] achieved similar results. They also generalize the sigmoid function to a large class of activation functions, showing that universal approximation is essentially implied by the network structure. Delalleau et al.[4] showed that there exists a family of functions which can be represented much more efficiently with deep networks than with shallow ones as well.

Since the development and success of deep neural networks recently, there have been much more works discussing the expressive power of neural networks theoretically. Depth efficiency is among the most typical results.

Other works turn to show deep networks' ability to approximate a wide range of functions.  For example,
Liang et al.[11] showed that in order to approximate a function which is $\Theta(\log \frac{1}{\epsilon})$-order derivable with $\epsilon$ error universally, a deep network with $O(\log \frac{1}{\epsilon})$ layers and $O(\mathrm{poly} \log \frac{1}{\epsilon})$ weights can do but $\Omega(\mathrm{poly} \frac{1}{\epsilon})$ weights will be required if there is only $o(\log \frac{1}{\epsilon})$ layers. Yarotsky [15] showed that $C^n$-functions on $\mathbb{R}^d$ with a bounded domain can be approximated with $\epsilon$ error universally by a ReLU network with $O(\log \frac{1}{\epsilon})$ layers and $O((\frac{1}{\epsilon})^{\frac{d}{n}}\log \frac{1}{\epsilon})$ weights. In addition, for results based on classic theories, Harvey et al.[7] provided a nearly-tight bound for VC-dimension of neural networks, that the VC-dimension for a network with $W$ weights and $L$ layers will have a $O(WL \log W)$ but $\Omega(WL \log \frac{W}{L})$ VC-dimension. Also, there are several works arguing for width's importance from other aspects, for example, Nguyen et al.[16] shows if a deep architecture is at the same time sufficiently wide at one hidden layer then it has a well-behaved loss surface in the sense that almost every critical point with full rank weight matrices is a global minimum from the view of optimization.

The remainder of the paper is organized as follows. In section 2 we introduce some background knowledge needed in this article.  In section 3 we present our main result -- the Width-Bounded Universal Approximation Theorem; besides, we show two comparing results related to the theorem.  Then in section 4 we turn to explore quantitatively the role of width for the expressive power of neural networks.  Finally, section 5 concludes.  All proofs can be found in the Appendix and we give proof sketch in main text as well.

\section{Preliminaries}

We begin by presenting basic definitions that will be used throughout the paper.  A neural network is a directed computation graph, where the nodes are computation units and the edges describe the connection pattern among the nodes.  Each node receives as input a weighted sum of activations flowed through the edges, applies some kind of activation function, and releases the output via the edges to other nodes.  Neural networks are often organized in layers, so that nodes only receive signals from the previous layer and only release signals to the next layer.  A fully-connected neural network is a layered neural network where there exists a connection between every two nodes in adjacent layers.  In this paper, we will study the fully-connected ReLU network, which is a fully-connected neural network with Rectifier Linear Unit (ReLU) activation functions.  The ReLU function $\mathrm{ReLU} \colon \mathbb{R} \to \mathbb{R}$ can be formally defined as

\begin{equation}
\mathrm{ReLU}(x) = \max (x, 0)
\end{equation}

The architecture of neural networks often specified by the width and the depth of the networks.  The depth $h$ of a network is defined as its number of layers (including output layer but excluding input layer); while the width $d_m$ of a network is defined to be the maximal number of nodes in a layer.  The number of input nodes, i.e. the input dimension, is denoted as $n$.

In this paper we study the expressive power of neural networks.  The expressive power describes neural networks' ability to approximate functions.  We focus on Lebesgue-integrable functions.  A Lebesgue-integrable function $f \colon \mathbb{R}^n \to \mathbb{R}$ is a Lebesgue-measurable function satisfying

\begin{equation}
\int_{\mathbb{R}^n} \vert f(x) \vert \mathrm{d} x < \infty
\end{equation}

which includes continuous functions, including functions such as the $\mathrm{sgn}$ function.  Because we deal with Lebesgue-integrable functions, we adopt $L^1$ distance as a measure of approximation error, different from $L^{\infty}$ distance used by some previous works which consider continuous functions.

\section{Width-bounded ReLU Networks as Universal Approximator}

In this section we consider universal approximation with width-bounded ReLU networks. The following theorem is the main result of this section.

\begin{theorem}[Universal Approximation Theorem for Width-Bounded ReLU Networks]
\label{th-ua}
For any Lebesgue-integrable function $f \colon \mathbb{R}^n\to \mathbb{R}$ and any $\epsilon>0$, there exists a fully-connected ReLU network $\mathscr{A}$ with width $d_m \le n + 4$, such that the function $F_{\mathscr{A}}$ represented by this network satisfies
\begin{equation}
\int_{\mathbb{R}^n} \vert f(x)-F_{\mathscr{A}}(x)\vert \mathrm{d} x < \epsilon.
\end{equation}
\end{theorem}

The proof of this theorem is lengthy and is deferred to the supplementary material. Here we provide an informal description of the high level idea.

For any Lebesgue integrable function and any predefined approximation accuracy, we explicitly construct a width-$(n+4)$ ReLU network so that it can approximate the function to the given accuracy. The network is a concatenation of a series of blocks. Each block satisfies the following properties:

1) It is a depth-$(4n+1)$ width-$(n+4)$ ReLU network.

2) It can approximate any Lebesgue integrable function which is uniformly zero outside a cube with length $\delta$ to a high accuracy;

3) It can store the output of the previous block, i.e., the approximation of other Lebesgue integrable functions on different cubes;

4) It can sum up its current approximation and the memory of the previous approximations.

It is not difficult to see that the construction of the whole network is completed once we build the blocks. We illustrate such a block in Figure 1 . In this block, each layer has $n+4$ neurons. Each rectangle in Figure 1 represents a neuron, and the symbols in the rectangle describes the output of that neuron as a function of the block. Among the $n+4$ neurons, $n$ neurons simply transfer the input coordinates. For the other $4$ neurons, $2$ neurons store the approximation fulfilled by previous blocks. The other $2$ neurons help to do the approximation on the current cube. The topology of the block is rather simple. It is very sparse, each neuron connects to at most $2$ neurons in the next layer.

The proof is just to verify the construction illustrated in Figure 1 is correct. Because of the space limit, we defer all the details to the supplementary materials.

\begin{figure}[!h]
\centering
\includegraphics[width=1\textwidth]{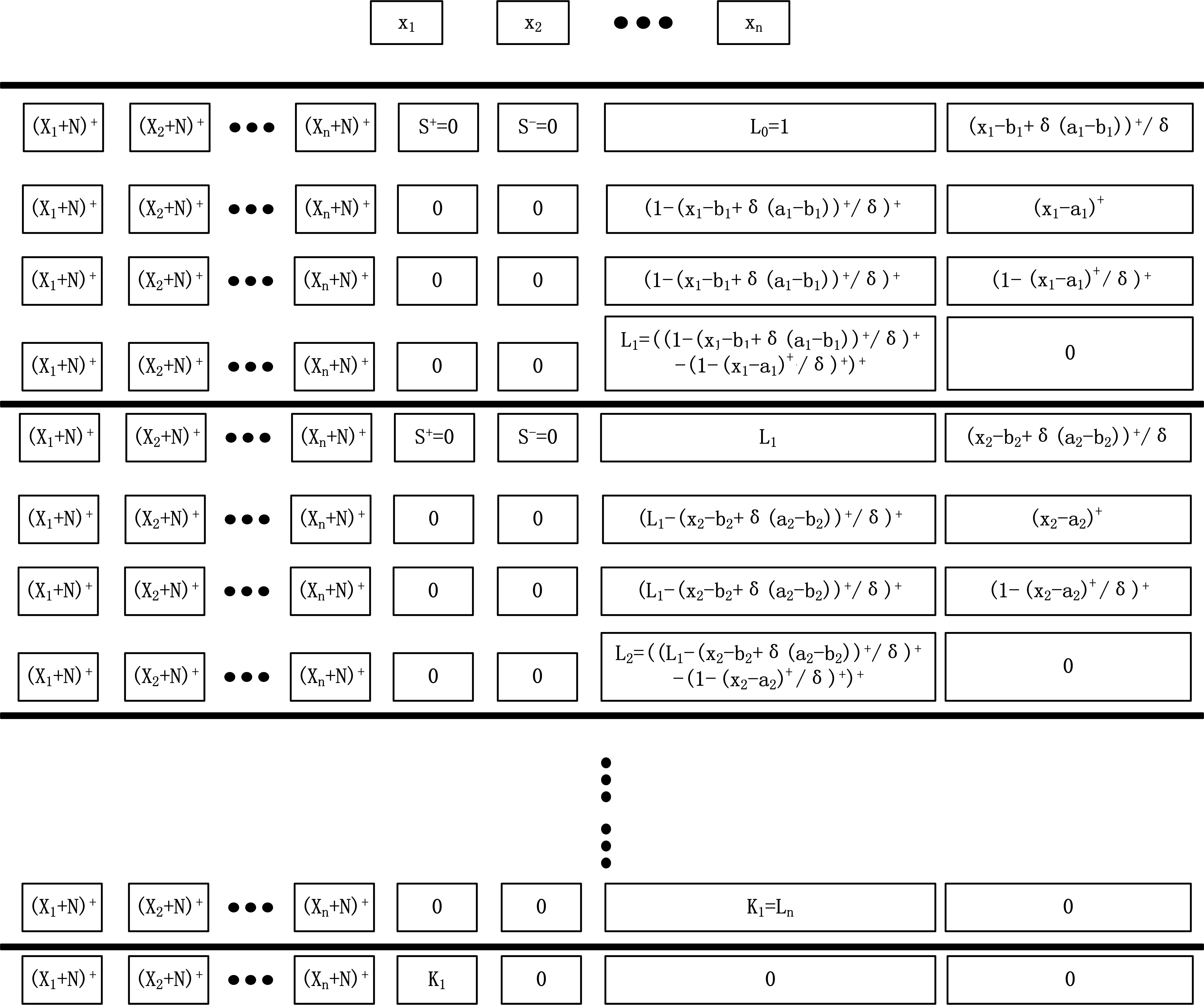}
\caption{One block to simulate the indicator function on $[a_{1},b_{1}]\times[a_{2},b_{2}]\times \dots \times [a_{n},b_{n}]$. For $k$ from $1$ to $n$, we "chop" two sides in the $kth$ dimension, and for every $k$ the "chopping" process is completed within a 4-layer sub-network as we show in Figure 1. It is stored in the (n+3)th node as $L_n$in the last layer of $\mathscr{A}$. We then use a single layer to record it in the (n+1)th or the (n+2)th node, and reset the last two nodes to zero. Now the network is ready to simulate another (n+1)-dimensional cube.}\label{fig:digit}
\end{figure}

Theorem 1 can be regarded as a dual version of the classical universal approximation theorem, which proves that depth-bounded networks are universal approximator. If we ignore the size of the network, both depth and width themselves are efficient for universal approximation. At the technical level however, there are a few differences between the two universal approximation theorems. The classical depth-bounded theorem considers continuous function on a compact domain and use $L^{\infty}$ distance; Our width-bounded theorem instead deals with Lebesgue-integrable functions on the whole Euclidean space and therefore use $L^1$ distance.

Theorem 1 implies that there is a phase transition for the expressive power of ReLU networks as the width of the network varies across $n$, the input dimension. It is not difficult to see that if the width is much smaller than $n$, then the expressive power of the network must be very weak. Formally, we have the following two results.

\begin{theorem}
\label{th-pt}
For any Lebesgue-integrable function $f \colon \mathbb{R}^n \to \mathbb{R}$ satisfying that $\{x:f(x) \neq 0 \}$ is a positive measure set in Lebesgue measure, and any function $F_{\mathscr{A}}$ represented by a fully-connected ReLU network ${\mathscr{A}}$ with width $d_m \le n$, the following equation holds:
\begin{equation}
\int_{\mathbb{R}^n} \vert f(x)-F_{\mathscr{A}}(x) \vert \mathrm{d} x = +\infty \ or \int_{\mathbb{R}^n} \vert f(x) \vert \mathrm{d} x.
\end{equation}
\end{theorem}

Theorem 2 says that even the width equals $n$, the approximation ability of the ReLU network is still weak, at least on the Euclidean space $\mathbb{R}^n$. If we restrict the function on a bounded set, we can still prove the following theorem.

\begin{theorem}
\label{co-pt}
For any continuous function $f \colon [-1,1]^n \to \mathbb{R}$ which is not constant
along any direction, there exists a universal $\epsilon^* >0$ such that for any function $F_A$ represented by a fully-connected ReLU network with width $d_{m} \leq n-1$, the $L^1$ distance between $f$ and $F_A$ is at least $\epsilon^*$:
\begin{equation}
\int_{[-1,1]^n} \vert f(x)-F_A(x) \vert \mathrm{d}x \geq \epsilon^*.
\end{equation}
Then it's a direct comparison with Theorem 1 since in Theorem 1 the $L^1$ distance can be arbitrarily small.
\end{theorem}

The main idea of the two theorems is grabbing the disadvantage brought by the insufficiency of dimension. If the corresponding first layer values of two different input points are the same, the output will be the same as well. When the ReLU network's width is not larger than the input layer's width, we can find a ray for "most" points such that the ray passes the point and the corresponding first layer values on the ray are the same. It is like a dimension reduction caused by insufficiency of width. Utilizing this weakness of thin network, we can finally prove the theorem.

\section{Width Efficiency vs. Depth Efficiency}

Going deeper and deeper has been a trend in recent years, starting from the 8-layer AlexNet [10], the 19-layer VGG [12], the 22-layer GoogLeNet [13], and finally to the 152-layer and 1001-layer ResNets [8].  The superiority of a larger depth has been extensively shown in the applications of many areas. For example, ResNet has largely advanced the state-of-the-art performance in computer vision related fields, which is claimed solely due to the extremely deep representations. Despite of the great practical success, theories of the role of depth are still limited.

Theoretical understanding of the strength of depth starts from analyzing the depth efficiency, by proving the existence of deep neural networks that cannot be realized by any shallow network whose size is exponentially larger. However, we argue that even for a comprehensive understanding of the depth itself, one needs to study the dual problem of width efficiency: Because, if we switch the role of depth and width in the depth efficiency theorems and the resulting statements remain true, then width would have the same power as depth for the expressiveness, at least in theory. It is worth noting that a priori, depth efficiency theorems do not imply anything about the validity of width efficiency.

In this section, we study the width efficiency of ReLU networks quantitatively.

\begin{theorem}
\label{th-pl}
Let $n$ be the input dimension. For any integer $k \ge n+4 $, there exists  $F_{\mathscr{A}} \colon \mathbb{R}^n \to \mathbb{R}$ represented by a ReLU neural network $\mathscr{A}$ with width $d_m = 2k^2$ and depth $h = 3$, such that for any constant $b>0$, there exists $\epsilon>0$ and for any function $F_{\mathscr{B}} \colon \mathbb{R}^n \to \mathbb{R}$ represented by ReLU neural network $\mathscr{B}$ whose parameters are bounded in $[-b,b ]$ with width $d_m \le k^{3/2}$ and depth $h \le k+2$, the following inequality holds:
\begin{equation}
\int_{\mathbb{R}^n} \left( F_{\mathscr{A}} - F_{\mathscr{B}} \right) ^ 2 \mathrm{d}x \ge \epsilon.
\end{equation}
\end{theorem}

Theorem 4 states that there are networks such that reducing width requires increasing in the size to compensate, which is similar to that of depth qualitatively. However, at the quantitative level, this theorem is very different to the depth efficiency theorems in [14][5][2]. Depth efficiency enjoys exponential lower bound, while for width Theorem 4 is a polynomial lower bound. Of course if a corresponding polynomial upper bound can be proven, we can say depth plays a more important role in efficiency, but such a polynomial lower bound still means that depth is not strictly stronger than width in efficiency ,sometimes it costs depth super-linear more nodes than width.

This raises a natural question: Can we improve the polynomial lower bound? There are at least two possibilities.

1) Width efficiency has \emph{exponential lower bound}. To be concrete, there are wide networks that cannot be approximated by any narrow networks whose size is no more than an exponential bound.

2) Width efficiency has \emph{polynomial upper bound}. Every wide network can be approximated by a narrow network whose size increase is no more than a polynomial.

Exponential lower bound and polynomial upper bound have completely different implications. If exponential lower bound is true, then width and depth have the same strength for the expressiveness, at least in theory. If the polynomial upper bound is true, then depth plays a significantly stronger role for the expressive power of ReLU networks.

Currently, neither the exponential lower bound nor the polynomial upper bound seems within the reach. We pose it as a formal open problem.

\subsection{Experiments}

We further conduct extensive experiments to provide some insights about the upper bound of such an approximation.  To this end, we study a series of network architectures with varied width.  For each network architecture, we randomly sample the parameters, which, together with the architecture, represent the function that we would like narrower networks to approximate.  The approximation error is empirically calculated as the mean square error between the target function and the approximator function evaluated on a series of uniformly placed inputs.  For simplicity and clearity,  we refer to the network architectures that will represent the target functions when assigned parameters as target networks, and the corresponding network architectures for approximator functions as approximator networks.

To be detailed, the target networks are fully-connected ReLU networks of input dimension $n$, output dimension $1$, width $2 k^ 2$ and depth $3$, for $n = 1, 2$ and $k = 3, 4, 5$.  For each of these networks, we sample weight parameters according to standard normal distribution, and bias parameters according to uniform distribution over $[-1, 1)$.  The network and the sampled parameters will collectively represent a target function that we use a narrow approximator network of width $3 k^{3/2}$ and depth $k+2$ to approximate, with a corresponding $k$.  The architectures are designed in accordance to Theorem 4 -- we aim to investigate whether such a lower bound is actually an upper bound.  In order to empirically calculate the approximation error, $20000$ uniformly placed inputs from $[-1, 1)^n$ for $n = 1$ and $40000$ such inputs for $n = 2$ are evaluated by the target function and the approximator function respectively, and the mean square error is reported.  For each target network, we repeat the parameter-sampling process $50$ times and report the mean square error in the worst and average case.

We adopt the standard supervised learning approach to search in the parameter space of the approximator network to find the best approximator function.  Specifically, half of all the test inputs from $[-1, 1)^n$ and the corresponding values evaluated by target function constitute the training set.  The training set is used to train approximator network with a mini-batch AdaDelta optimizer and learning rate $1.0$.  The parameters of approximator network are randomly initialized according to [8]. The training process proceeds $100$ epoches for $n = 1$ and $200$ epoches for $n = 2$; the best approximator function is recorded.

Table 1 lists the results.  Figure 2 illustrates the comparison of an example target function and the corresponding approximator function for $n = 1$ and $k = 5$.  Note that the target function values vary with a scale $\sim 10$ in the given domain, so the (absolute) mean square error is indeed a rational measure of the approximation error.  It is shown that the approximation error is indeed very small, for the target networks and approximator networks we study. From Figure 2 we can see that the approximation function is so close to the target function that we have to enlarge a local region to better display the difference. Since the architectures of both the target networks and approximator networks are determined according to Theorem 4, where the depth of approximator networks are in a polynomial scale with respect to that of target networks, the empirical results show an indication that a polynomial larger depth may be sufficient for a narrow network to approximate a wide network.

\begin{table}
  \caption{Empirical study results.  $n$ denotes the input dimension, $k$ is defined in Theorem 4; the width/depth for both target network and approximator network are determined in accordance to Theorem 4.  We report mean square error in the worst and average case over $50$ runs of randomly sampled parameters for target network.}
  \label{tb}
  \centering
  \begin{tabular}{c c c c c c c c}
    \toprule
    \multirow{2}{*}{$n$} & \multirow{2}{*}{$k$} & \multicolumn{2}{c}{target network} & \multicolumn{2}{c}{approximator network} & \multirow{2}{*}{worst case error} & \multirow{2}{*}{average case error}\\  
    \cmidrule(lr){3-4} \cmidrule(lr){5-6}
     & & width & depth & width & depth & & \\
    \midrule
    1 & 3 & 18 & 3 & 16 & 5 & 0.002248 & 0.000345 \\
    1 & 4 & 36 & 3 & 24 & 6 & 0.003263 & 0.000892 \\
    1 & 5 & 50 & 3 & 34 & 7 & 0.005643 & 0.001296 \\
    2 & 3 & 18 & 3 & 16 & 5 & 0.008729 & 0.001990 \\
    2 & 4 & 36 & 3 & 24 & 6 & 0.018852 & 0.006251 \\
    2 & 5 & 50 & 3 & 34 & 7 & 0.030114 & 0.007984 \\
    \bottomrule
  \end{tabular}
\end{table}

\begin{figure}[h]
\centering
\includegraphics[width=0.6\textwidth]{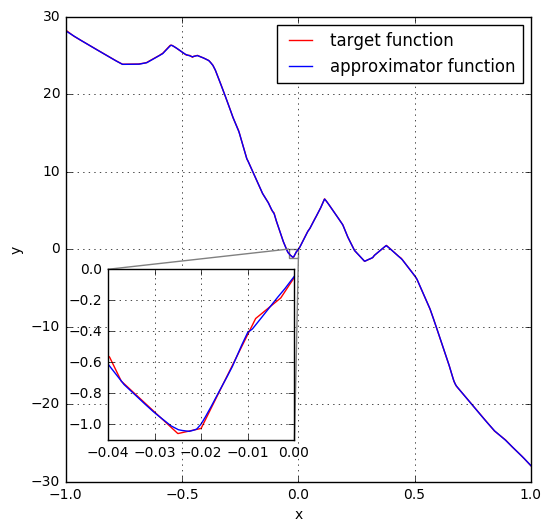}
\caption{Comparison of an example target function and the corresponding approximator function for $n = 1$ and $k = 5$. A local region is enlarged to better display the difference.}
\label{fig}
\end{figure}

\section{Conclusion}

In this paper, we analyze the expressive power of neural networks with a view from the \emph{width}, distinguished from many previous works which focus on the view from the \emph{depth}.  We establish the Universal Approximation Theorem for Width-Bounded ReLU Networks, in contrast with the well-known Universal Approximation Theorem, which studies depth-bounded networks. Our result demonstrate a phase transition with respect to expressive power when the width of a ReLU network of given input dimension varies. 

We also explore the role of width for the expressive power of neural networks: we prove that a wide network cannot be approximated by a narrow network unless with polynomial more nodes, which gives a lower bound of the number of nodes for approximation. We pose open problems on whether exponential lower bound or polynomial upper bound hold for the width efficiency, which we think is crucial on the way to a more thorough understanding of expressive power of neural networks. Experimental results support the polynomial upper bound and agree with our intuition and insights from the analysis.

The width and the depth are two key components in the design of a neural network architecture. Width and depth are both important and should be carefully tuned together for the best performance of neural networks, since the depth may determine the abstraction level but the width may influence the loss of information in the forwarding pass. A comprehensive understanding of the expressive power of neural networks requires looking from both views.

\section*{Acknowledgments}

This work was partially supported by National Basic Research Program of China (973 Program)
(grant no. 2015CB352502), NSFC (61573026), and the elite undergraduate training program of School of Mathematical Science in Peking University. We would like to thank the anonymous reviewers for their valuable comments on our paper.

\section*{References}

\small
[1] Andrew R Barron. Approximation and estimation bounds for artificial neural networks. Machine
Learning, 14(1):115–133, 1994.

[2] Nadav Cohen, Or Sharir, and Amnon Shashua. On the expressive power of deep learning: A
tensor analysis. In Conference on Learning Theory, pages 698–728, 2016.

[3] George Cybenko. Approximation by superpositions of a sigmoidal function. Mathematics of
Control, Signals, and Systems (MCSS), 2(4):303–314, 1989.

[4] Olivier Delalleau and Yoshua Bengio. Shallow vs. deep sum-product networks. In Advances in
Neural Information Processing Systems, pages 666–674, 2011.

[5] Ronen Eldan and Ohad Shamir. The power of depth for feedforward neural networks. In
Conference on Learning Theory, pages 907–940, 2016.

[6] Ken-Ichi Funahashi. On the approximate realization of continuous mappings by neural networks.
Neural networks, 2(3):183–192, 1989.

[7] Nick Harvey, Chris Liaw, and Abbas Mehrabian. Nearly-tight vc-dimension bounds for piecewise
linear neural networks. COLT 2017, 2017.

[8] Kaiming He, Xiangyu Zhang, Shaoqing Ren, and Jian Sun. Deep residual learning for image
recognition. In Proceedings of the IEEE Conference on Computer Vision and Pattern
Recognition, pages 770–778, 2016.

[9] Kurt Hornik, Maxwell Stinchcombe, and Halbert White. Multilayer feedforward networks are
universal approximators. Neural networks, 2(5):359–366, 1989.

[10] Alex Krizhevsky, Ilya Sutskever, and Geoffrey E Hinton. Imagenet classification with deep
convolutional neural networks. In Advances in neural information processing systems, pages
1097–1105, 2012.

[11] R. Srikant Shiyu Liang. Why deep neural networks for funtion approximation? ICLR 2017,
2017.

[12] Karen Simonyan and Andrew Zisserman. Very deep convolutional networks for large-scale
image recognition. CoRR, abs/1409.1556, 2014.

[13] Christian Szegedy, Wei Liu, Yangqing Jia, Pierre Sermanet, Scott E. Reed, Dragomir Anguelov,
Dumitru Erhan, Vincent Vanhoucke, and Andrew Rabinovich. Going deeper with convolutions.
CoRR, abs/1409.4842, 2014.

[14] Matus Telgarsky. Benefits of depth in neural networks. COLT 2016: 1517-1539, 2016.

[15] Dmitry Yarotsky. Error bounds for approximations with deep relu networks. arXiv preprint
arXiv:1610.01145, 2016.

[16] Quynh Nguyen and Matthias Hein. The loss surface of deep and wide neural networks. In
Doina Precup and Yee Whye Teh, editors, Proceedings of the 34th International Conference on
Machine Learning, volume 70 of Proceedings of Machine Learning Research, pages 2603–2612,
International Convention Centre, Sydney, Australia, 06–11 Aug 2017. PMLR.
\newpage

\appendix
\section{Appendix}
\subsection{Proof of Theorem 1}
\begin{proof}
We prove this theorem by constructing a network architecture which can approximate any Lesbegue-integrable function w.r.t $L^1$ distance. We will firstly illustrate that $f$ can be approximated by finite weighted sum of indicator functions on n-dimensional cubes. Then we will show how a ReLU network approximate an indicator function on an n-dimensional cube. Finally we will show that ReLU network can "store" the quantities and sum them up.

Assume $x=(x_{1},\dots,x_{n})$ is the input.
Since $f$ is L-integrable, for any $\epsilon>0$, there exists $N>0$ which satisfies \[\int_{\cup_{i=1}^n |x_{i}|\ge N} \vert f\vert dx< \frac{\epsilon}{2}\]
For simplication, the following symbols are introduced.
\[E \triangleq [-N,N]^{n}\]
\[ f_{1}(x)\triangleq\begin{cases}
max\{f,0\} & x\in E\\
0 & x\notin E
\end{cases} \]
\[ f_{2}(x)\triangleq\begin{cases}
max\{-f,0\} & x\in E\\
0 & x\notin E
\end{cases} \]
\[C \triangleq \int_{R^n}\vert f\vert d\vec{x}\]
\[V_{E}^{1} \triangleq \{(x,y) \vert x \in E, 0<y<f_{1}(x))\}\]
\[V_{E}^{2} \triangleq \{(x,y) \vert x \in E, 0<y<f_{2}(x))\}\]
Then we have
\begin{align}\label{eq1}
\int_{R^n} |f-(f_{1}-f_{2})|dx<\frac{\epsilon}{2} 
\end{align}
$f_1$ denotes the positive part of$f$, while $f_2$ denotes the negative part. $V_{E}^{i}$ is the space between $f_{i}$ and $y=0$ in $E$, i=1,2.

For i=1,2, since $V_{E}^{i}$ is measurable, there exists a Lebesgue cover of $V_{E}^{i}$ consisting finite (n+1)-dimensional cubes $J_{j,i}$, satisfying
\begin{align}\label{eq2}
m (V_{E}^{i}\bigtriangleup \bigcup_j J_{j,i}) < \frac{\epsilon}{8}
\end{align}. We assume the number of $J_{j,i}s$ is $n_i$.
Here and below $m(\cdot)$ denotes Lebesgue measure.

For any (n+1)-dimensional cube $J_{j,i}$, we assume
$$J_{j,i}=[a_{1,j,i},a_{1,j,i}+b_{1,j,i}]\times [a_{2,j,i},a_{2,j,i}+b_{2,j,i}]\times \dots \times [a_{n+1,j,i},a_{n+1,j,i}+b_{n+1,j,i}]$$
$$X_{j,i}=[a_{1,j,i},a_{1,j,i}+b_{1,j,i}]\times [a_{2,j,i},a_{2,j,i}+b_{2,j,i}]\times \dots \times [a_{n,j,i},a_{n,j,i}+b_{n,j,i}]$$
Note that each $J_{j,i}$ corresponds to an indicator function. we define
\[ \phi_{j,i}(x)=\begin{cases}
1& x\in X_{j,i}\\
0 & x\notin X_{j,i}
\end{cases} \]
Based on inequality \eqref{eq2}, we have
\begin{align}\label{eq3}
\int_{E}|f_{i}-\sum_{j=1}^{n_{i}}b_{n+1,j,i}\phi_{j,i}|dx<\frac{\epsilon}{8}
\end{align}
From \eqref{eq1} and \eqref{eq3}, we can prove that f can be approximated by finite weighted sum of indicator function on n-dimensional cubes. Also we have
\begin{align}
\sum_{i=1}^{2}\int_{E}|\sum_{j=1}^{n_{i}}b_{n+1,j,i}\phi_{j,i}|dx&=\sum_{i=1}^{2}\sum_{j=1}^{n_{i}}\int_{E}b_{n+1,j,i}\phi_{j,i}dx\\
&<C+\frac{3\epsilon}{4}
\end{align}
Then we will show how to use ReLU network to approximate such a function.\\
We wish to find functions $\varphi_{j,i}$, satisfying
\begin{align}
\int_{X_{j,i}}|\phi_{j,i}-\varphi_{j,i}|dx &<\frac{\epsilon}{4(C+\frac{3\epsilon}{4})}\int_{E}|\phi_{j,i}|dx\\ \label{eq5}
&=\frac{\epsilon}{4C+3\epsilon}\int_{E}|\phi_{j,i}|dx
\end{align}

For any I $\in$ \{$\phi_{j,i}$\}, we assume
\[ I=\begin{cases}
1& x\in X\\
0 & x\notin X
\end{cases} \]
Here, $$X=[a_{1},b_{1}]\times[a_{2},b_{2}]\times \dots \times [a_{n},b_{n}]$$
Apparently,
$$a_{j},b_{j}\in [-N,N],j=1,2,\dots,n$$
Next we will construct a network $\mathscr{A}$ to produce a function J, satisfying
\begin{align}
\int_{E}|I-J|dx&<\frac{\epsilon}{4C+3\epsilon}\int_{E}Idx\\
&=\frac{\epsilon}{4C+3\epsilon}\prod_{i=1}^{n}(b_i-a_i) \label{eq4}
\end{align}
We define some notations here. We denote the network by $\mathscr{A}$, the function represented by the whole network by $F_{\mathscr{A}}$, the function represented by the $kth$ layer of the network by $F_{k,\mathscr{A}}$, the function represented by the $jth$ node in the $kth$ layer by $F_{k,j,\mathscr{A}}$, the function represented by the first $k$ layers of the network after being ReLUed by $R_{k,\mathscr{A}}$. The function represented by the $jth$ node in the $kth$ layer after ReLUed is $R_{k,j,\mathscr{A}}$. Here, without loss of generality, $R_{0,\mathscr{A}}$ denotes the input layer. The weight matrix is denoted by $A$ and the offset vector by $u$. The depth is denoted by h.

For any $\delta>0,k=1,2,\dots,n$, we can design a ReLU network $\mathscr{A}_k$ satisfying following conditions:\\
(1)The width of each layer of $\mathscr{A}_k$ is n+4.\\
(2)The depth of $\mathscr{A}$ is 3.\\
(3)for i=0,1,2,3, j=1,2,\dots,n, $R_{i,j,\mathscr{A}_k}=(x_{i}+N)^{+}$\\
(4)for j=n+1,n+2, all the weights related to $R_{i,j,\mathscr{A}_k}$ are 0.\\
(5)$R_{1,n+3,\mathscr{A}_k}$ is a function of x such that
\begin{itemize}
\item $0\leq R_{1,n+3,\mathscr{A}_k}(x)\leq 1$ for any x
\item $R_{1,n+3,\mathscr{A}_k}(x)=0$ if $(x_{1},\dots,x_{k-1})\notin [a_{1},b_{1}]\times \dots \times [a_{k-1},b_{k-1}]$
\item $R_{1,n+3,\mathscr{A}_k}(x)=1$ if $(x_{1},\dots,x_{k-1})\in [a_1+\delta(b_1-a_1),b_1-\delta(b_1-a_1)]\times \dots \times [a_{k-1}+\delta(b_{k-1}-a_{k-1}),b_{k-1}-\delta(b_{k-1}-a_{k-1})]$
\end{itemize}
(6) $R_{3,n+3,\mathscr{A}_k}$ is a function of x such that
\begin{itemize}
\item $0\leq R_{4,n+3,\mathscr{A}_k}(x)\leq 1$ for any x
\item $R_{4,n+3,\mathscr{A}_k}(x)=0$ if $(x_{1},\dots,x_{k})\notin [a_{1},b_{1}]\times \dots \times [a_{k},b_{k}]$
\item $R_{4,n+3,\mathscr{A}_k}(x)=1$ if $(x_{1},\dots,x_{k})\in [a_1+\delta(b_1-a_1),b_1-\delta(b_1-a_1)]\times \dots \times [a_{k}+\delta(b_{k}-a_{k}),b_{k}-\delta(b_{k}-a_{k})]$
\end{itemize}
We call this shallow ReLU network Single ReLU Unit(SRU). We will explain some details of SRU. The first n+2 nodes in each layer is "memory element" of SRU while the last two is the "computation element" of SRU. The main idea of SRU is to process the function $R_{0,n+3,\mathscr{A}_k}$ to get $R_{3,n+3,\mathscr{A}_k}$.\\

The main idea of this process is to "chop" the function and reduce the support set of the function. See Figure 1 for a simulation sample when $n=2$.
\begin{figure}[!h]
\centering
\includegraphics[width=0.85\textwidth]{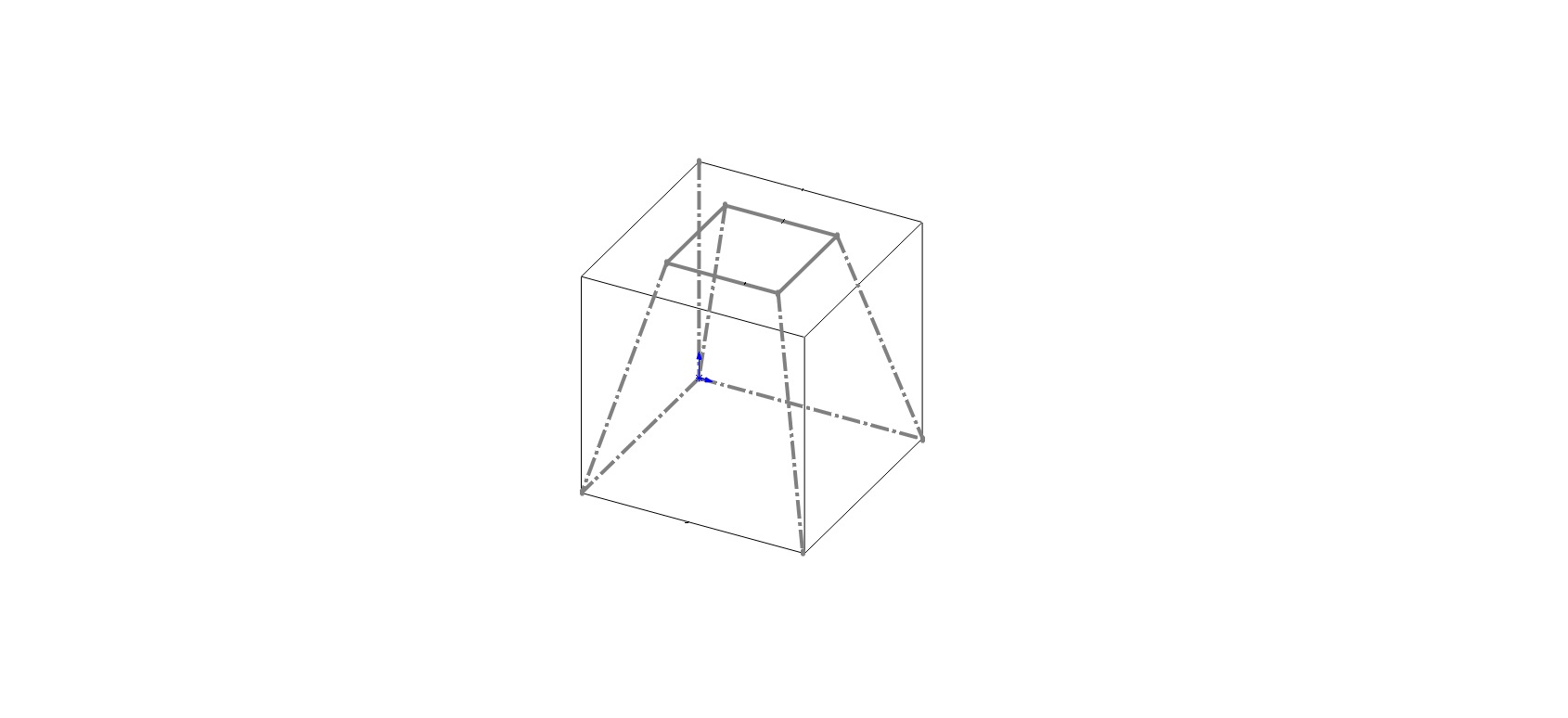}
\caption{cube $I$ and hyper-trapezoid $J$ inside $I$}\label{fig:digit}
\end{figure}

Denote
\[\mathscr{A}=\mathscr{A}_n \comp \mathscr{A}_{n-1} \comp \cdots \comp \mathscr{A}_{1}\]

We will show that, for any $\delta>0$, $J = \mathscr{A}(x_1,x_2,\cdots,x_n)$ can produce exatly the same shape as the hyper-trapezoid inscribed in cube $I$ in Figure 1. For simplicity, define $\mathscr{B}_k=\mathscr{A}_k \comp \mathscr{A}_{k-1} \comp \cdots \comp \mathscr{A}_{1}$, here $k=1,2,\cdots,n$.\\
Examine $\mathscr{B}_1$. The input layer is identity function in every dimension.
\[R_{0,j,\mathscr{B}_1}=x_j\]
For simplicity, define $f^+ = ReLU(f)$. The first hidden layer retains the information of the input layer.
\[ R_{1,j,\mathscr{B}_1}=\begin{cases}
(x_j+N)^+& j=1,2,\cdots,n\\
0 & j=n+1,n+2\\
1 & j=n+3\\
(x_1-b_1+\delta(b_1-a_1))^+ & j=n+4
\end{cases} \]
The first n nodes remain unchanged thorough out the whole network $\mathscr{A}$, which are used to record the information of the input layer.The $(n+1)$ and $(n+2)$th node are reserved for the positive and negetive part of the whole target function respectively. In fact, the whole network $\mathscr{A}$ is constructed to simulate a single indicator function $I$, if the function $I$ is positive, then we will store the simulation result $J$ into the $(n+1)$th node. Otherwise, $J$ will be stored into $(n+2)$th node. By adding up those simulation results in these two nodes, we can get a simulation of $\sum_{j=1}^{n_{i}}(-1)^{i+1}b_{n+1,j,i}\phi_{j,i}$ , and thus simulates the target function.
We list the result in second,third and fourth layer below.
\[ R_{2,j,\mathscr{B}_1}=\begin{cases}
(x_j+N)^+& j=1,2,\cdots,n\\
0 & j=n+1,n+2\\
(1-\frac{(x_1-b_1+\delta(b_1-a_1))^+}{\delta})^+ & j=n+3\\
(x_1-a_1)^+ & j=n+4
\end{cases} \]
\[ R_{3,j,\mathscr{B}_1}=\begin{cases}
(x_j+N)^+& j=1,2,\cdots,n\\
0 & j=n+1,n+2\\
(1-\frac{(x_1-b_1+\delta(b_1-a_1))^+}{\delta})^+ & j=n+3\\
(1-\frac{(x_1-a_1)^+}{\delta})^+ & j=n+4
\end{cases} \]
\[ R_{4,j,\mathscr{B}_1}=\begin{cases}
(x_j+N)^+& j=1,2,\cdots,n\\
0 & j=n+1,n+2\\
L_1 = ((1-\frac{(x_1-b_1+\delta(b_1-a_1))^+}{\delta})^+ - (1-\frac{(x_1-a_1)^+}{\delta})^+)^+& j=n+3\\
0 & j=n+4
\end{cases} \]

For simplicity, denote $L_k = R_{4,j,\mathscr{B}_k}$.The network $\mathscr{A}_k \quad (k=2,\cdots,n)$ is similar to the case of $k=1$.The input layer is the final layer in $\mathscr{B}_{k-1}$.
\[ R_{1,j,\mathscr{B}_k}=\begin{cases}
(x_j+N)^+& j=1,2,\cdots,n\\
0 & j=n+1,n+2\\
L_{k-1} & j=n+3\\
(x_k-b_k+\delta(b_k-a_k))^+ & j=n+4
\end{cases} \]
\[ R_{2,j,\mathscr{B}_k}=\begin{cases}
(x_j+N)^+& j=1,2,\cdots,n\\
0 & j=n+1,n+2\\
(1-\frac{(x_k-b_k+\delta(b_k-a_k))^+}{\delta})^+ & j=n+3\\
(x_k-a_k)^+ & j=n+4
\end{cases} \]
\[ R_{3,j,\mathscr{B}_k}=\begin{cases}
(x_j+N)^+& j=1,2,\cdots,n\\
0 & j=n+1,n+2\\
(1-\frac{(x_k-b_k+\delta(b_k-a_k))^+}{\delta})^+ & j=n+3\\
(1-\frac{(x_k-a_k)^+}{\delta})^+ & j=n+4
\end{cases} \]
\[ R_{4,j,\mathscr{B}_k}=\begin{cases}
(x_j+N)^+& j=1,2,\cdots,n\\
0 & j=n+1,n+2\\
L_k = (\frac{(x_k-b_k+\delta(b_k-a_k))^+}{\delta} - (1-\frac{(x_k-a_k)^+}{\delta})^+& j=n+3\\
0 & j=n+4
\end{cases} \]

For each k, we "chop" two sides in the kth dimension. Finally, we get the shape J in Figure 3.It is stored in the (n+3)th node as $L_n$in the last layer of $\mathscr{A}$. We then use a single layer to record it in the (n+1)th or the (n+2)th node, and reset the last two nodes to zero. Now the network is ready to simulate another (n+1)-dimensional cube. The whole construction process is shown in Figure 4.

\begin{figure}[!h]
\centering
\includegraphics[width=1\textwidth]{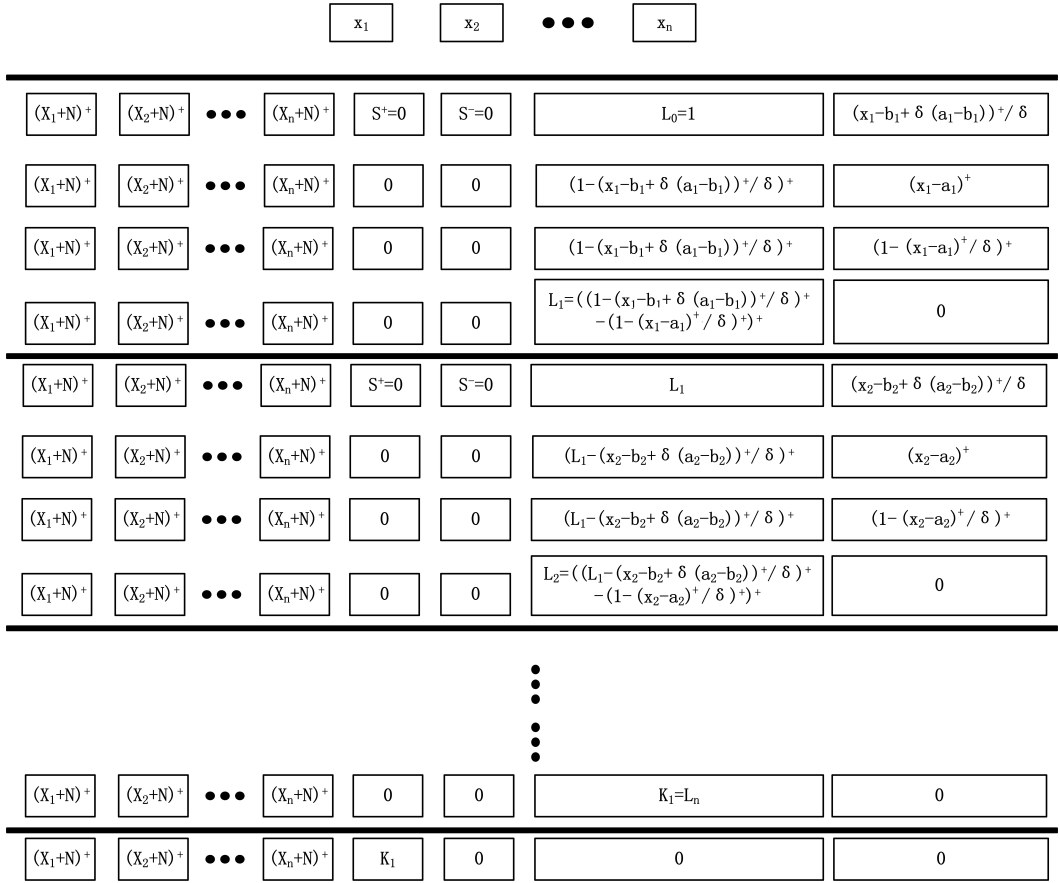}
\caption{The whole process to simulate a cube;every four layers are used to reshape one dimension of the cube(seperated by thick lines)}\label{fig:digit}
\end{figure}

Using this construction, we can simulate $I$ by $J$, which is produced by network $\mathscr{A}$. Note that, as $\delta$ approaches 0, the simulation error w.r.t $L_1$ distance converges to 0.

Next we will find a value of $\delta$ to fit the need of our proof. See Figure 3. The side length of small square on the top surface is $1-2\delta$ as the side length of the top surface. We will select a suitable $\delta>0$, satisfying $\int_{X}|I-J|d{x}<\frac{\epsilon}{4C+3\epsilon}\int_{E}|I|dx$.\\
Denote
\[X_0=[a_1+\delta(b_1-a_1),b_1-\delta(b_1-a_1)]\times \dots \times [a_{n}+\delta(b_{n}-a_{n}),b_{n}-\delta(b_{n}-a_{n})]\]
Notice that $I-J=0$ on $X_0$, and the maximum value of $I-J$ on $X$ is 1. Thus,
\begin{align}
\int_{X}|I-J|dx &<\int_{X}\textbf{1}_{x \in X\setminus X_{0}}dx\\ 
&=(1-(1-2\delta)^n)\prod_{i=1}^{n}(b_i-a_i) \label{eq6}
\end{align}
Compared with \eqref{eq4}, we set
\begin{align}
\delta = \frac{1-(1-\frac{\epsilon}{4C+3\epsilon})^{\frac{1}{n}}}{2}
\end{align}
Then we have
\begin{align}
\int_{X}|I-J|dx <\frac{\epsilon}{4C+3\epsilon}\prod_{i=1}^{n}(b_i-a_i)
\end{align}

Satisfies
\[\int_{X}|I-J|d{x}<\frac{\epsilon}{4C+3\epsilon}\int_{E}|I|dx\]
Thus, for $i=1,2;j=1,2,\cdots,n_i$, $\phi_{j,i}$ can be approximated by network function $\mu_{j,i}$. Satisfies
\[\int_{E}|\phi_{j,i}-\varphi_{j,i}|d{x}<\frac{\epsilon}{4C+3\epsilon}\int_{E}\phi_{j,i}dx\]
Sum those equations up, combined with \eqref{eq5}, we have
\begin{align}
\sum_{i=1}^{2}\sum_{j=1}^{n_{i}}\int_{E}|(-1)^{i+1}b_{n+1,j,i}(\phi_{j,i}-\mu_{j,i})|d{x}&<\frac{\epsilon}{4C+3\epsilon}\sum_{i=1}^{2}\sum_{j=1}^{n_{i}}\int_{E}b_{n+1,j,i}\phi_{j,i}dx\\
&\le \frac{\epsilon}{4C+3\epsilon}*(C+\frac{3\epsilon}{4})\\
&= \frac{\epsilon}{4}
\end{align}
Thus, we have the approximation of cubes $J_{j,i}$. Next we show how to combine those approximation functions together by network. There are $n_1$ positive cubes, corresponding to $n_1$ positive functions $\mu_{i,1}$;$n_2$ negative cubes, correspond to $n_2$ negative functions $\mu_{j,2}$. The detailed network is shown in Figure 3.

\begin{figure}[!h]
\centering
\includegraphics[width=0.7\textwidth]{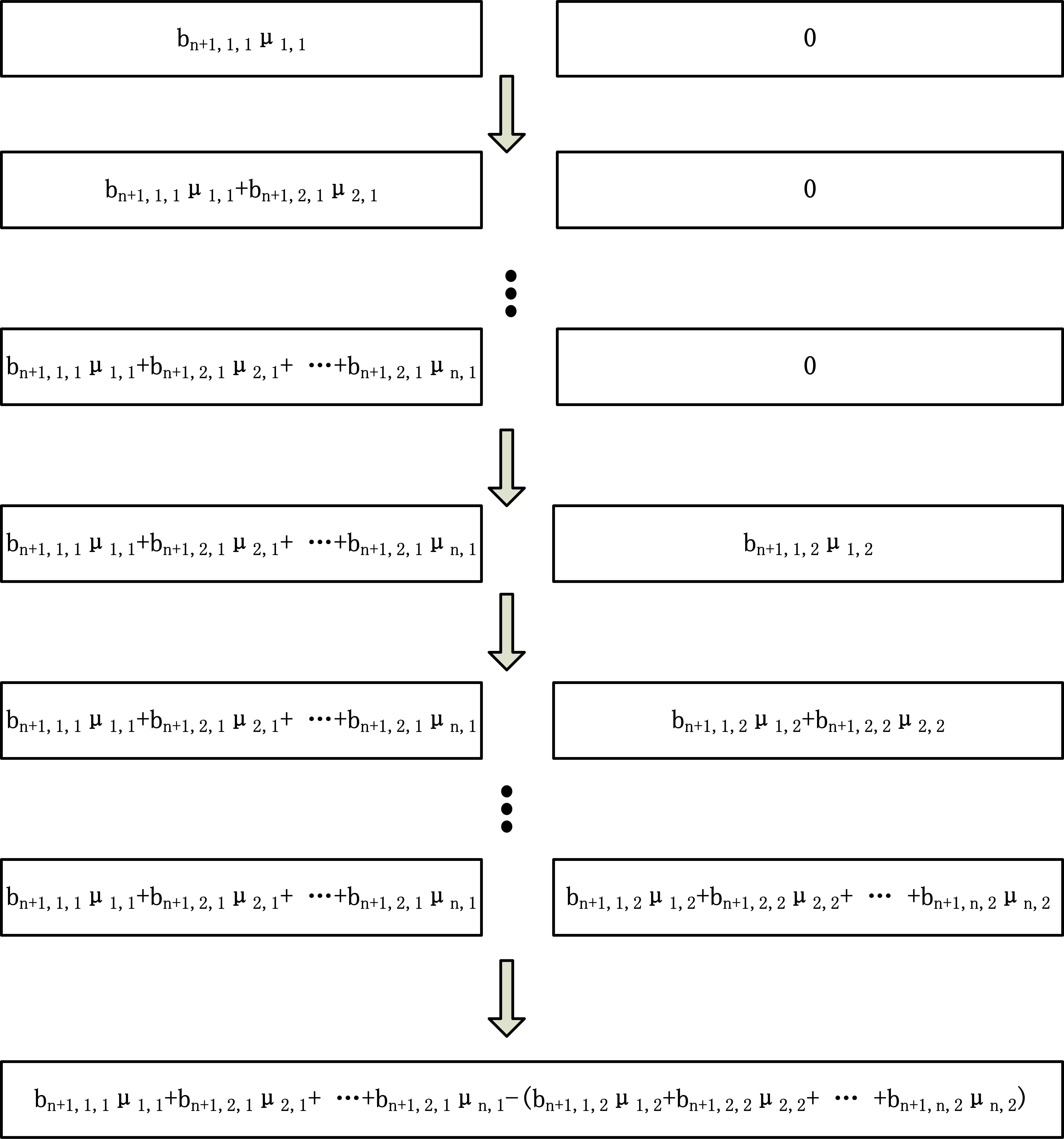}
\caption{The final process to simulate target function;every shown layer is the (n+1) and (n+2)th node in the last layer in Figure 4, which represent the simulation of a single cube. This figure shows the process of adding those functions up to get the function we want. Notice that except for the output layer, every result is nonnegative in the process and is produced by RELU activator. For simplicity, we just omit the RELU mark in the graph. }\label{fig:digit}
\end{figure}

Finally, we have $g \triangleq \sum_{i=1}^{2}\sum_{j=1}^{n_{i}}(-1)^{i+1}b_{n+1,j,i}\mu_{j,i}d{x}$. $f_0$ is the result function produced by our designed network. Combined with \eqref{eq1},\eqref{eq3},\eqref{eq6}, we have
\begin{align}
&\quad \int_{R^n} |f-g|dx\\
&<\int_{R^n} |f-(f_1-f_2)|dx+\sum_{i=1}^{2}\int_{E}|f_{i}-\sum_{j=1}^{n_{i}}(-1)^{i+1}b_{n+1,j,i}\phi_{j,i}|dx \nonumber\\
&+\sum_{i=1}^{2}\sum_{j=1}^{n_{i}}\int_{E}|(-1)^{i+1}b_{n+1,j,i}(\phi_{j,i}-\mu_{j,i})|d{x}\\
&<\frac{\epsilon}{2}+2*\frac{\epsilon}{8}+\frac{\epsilon}{4}\\
&=\epsilon
\end{align}

Thus, $g$ is the function we need in the theorem.

\end{proof}

\subsection{Proof of Theorem 2}

The proof is long and complicated, so we firstly define some notations for convenience afterwards. We denote the network by $\mathscr{A}$, the function represented by the whole network by $F_{\mathscr{A}}$, the function represented by the $kth$ layer of the network by $F_{k,\mathscr{A}}$, the function represented by the $jth$ node in the $kth$ layer by $F_{k,j,\mathscr{A}}$, the function represented by the first $k$ layers of the network after being ReLUed by $R_{k,\mathscr{A}}$. Here, without loss of generality, $R_{0,\mathscr{A}}$ denotes the input layer. We define 
 \par
 Condition 1: $d_{m}=n$ and the widths of all the layers except the output layer are n.
 \par
   Obviously other cases where $d_{m}\leq n$ are just special cases of this setting. The weight matrix of each layer is denoted by $A_{d}$ and the offset vector by $u_{d}$ where d is the number of layer. The depth is denoted by h.\\
 Here we will introduce 2 definitions inspired by \emph{Benefits of depth in neural networks} (Telgarsky ,2016).
\par
\textbf{Definition 1}:\quad A set X $\subset \ R^{n}$ is a \emph{linear block} if there exist t linear functions $(q_{i})_{i=1}^{t}$, and m tuples $(U_{j},L_{j})_{j=1}^{m}$ where $U_{j}$ and $L_{j}$ are subsets of [t](where [t]:={$1,\dots,t$}), such that $\vec{x} \in \ $X is equivalent to
$$(\Pi_{i\in L_{j}}1[q_{i}(v)<0])(\Pi_{i\in U_{j}}1[q_{i}(v)\geq 0])=1$$
\par
\textbf{Definition 2}:\quad A function f:$R^{k}\to R$ is  $(t,\alpha,\beta)-sa((t,\alpha,\beta)-semi-algebraic)$ if there exist t polynomials $(q_{i})_{i=1}^{t}$ of degree $\leq \alpha$,  and m triples $(U_{j},L_{j},p_{j})_{j=1}^{m}$ where $U_{j}$ and $L_{j}$ are subsets of [t](where [t]:={$1,\dots,t$}) and $p_j$ is a polynomial of degree $\leq \beta$, such that
$$f(v)=\Sigma_{j=1}^{m}p_{j}(v)(\Pi_{i\in L_{j}}1[q_{i}(v)<0])(\Pi_{i\in U_{j}}1[q_{i}(v)\geq 0])$$\\
We can see Theorem 2 is a direct conclusion of Lemma 1 as follows:\\

\textbf{Lemma 1}:\quad Consider a function $F_{\mathscr{A}}$ represented by a relu neural network $\mathscr{A}$ where $d_m \leq n$, the following equation holds.
$$\int_{R^n}|F_{\mathscr{A}}(\vec{x})|d\vec{x}=0 \ or +\infty$$
\par
We define assumption 1 here.
\par
Assumption 1:
$$\int_{R^{n}}|F_{\mathscr{A}}(\vec{x})|d\vec{x}<+\infty $$
We will prove that if assumption 1 holds, $$\int_{R^{n}}|F_{\mathscr{A}}(\vec{x})|d\vec{x}=0$$, which is equivalent to Lemma 1.
To prove Lemma 1, we need Lemma 2. \\
\par
\textbf{Lemma 2}:\quad For any given $\mathscr{A}$ where assumption 1 and Condition 1 hold and any $k\in\{0,1,2,\dots,h-1\}$, there exists a \emph{linear block} $X_k$ which satisfies following conditions:
\par
$S_{1}(k)$:$X_k$ is convex.
\par
$S_{2}(k)$:For any $\vec{x}\notin X_k$, $F_{\mathscr{A}}(\vec{x})=0$
\par
$S_{3}(k)$:For any $\vec{x}$ in $B(X_{k})$, $F_{\mathscr{A}}(\vec{x})=0$, where $B(X_{k})$, the boundary set of $X_k$, is defined as $\{\vec{x}:for \ any \ \epsilon >0, \exists \vec{u}\in X_{k},\vec{v}\notin X_{k} s.t. ||\vec{u}-\vec{x}||<\epsilon,||\vec{v}-\vec{x}||<\epsilon,  \}$
\par
$S_{4}(k)$:There exists a matrix $H$ and a vector $\vec{b}$ such that $R_{k,\mathscr{A}}(\vec{x})=H\vec{x}+\vec{b}$ for $\vec{x}\in X_{k}$\\
\par
If Lemma 2 holds and assumption 1 holds, let $k=h-1$, $F_{\mathscr{A}}$ is a linear function on  its support set, a \emph{linear block}. It is not hard to prove Lemma 1 after that. However, the proof of Lemma 2 is difficult. Before getting into the detail, we'd like to make some remark. Our conclusion may seem strange at first since $F_{\mathscr{A}}$ is like a linear function. Note we derive all these conclusions under assumption 1. Our proof actually shows that assumption 1 does not hold in most cases and the expressive power of thin neural networks is weak.\\
Before proving Lemma 2, we need Lemma 3 as a preparation.
\par
Apparently, for any Relu neural network $\mathscr{A}$, there exists an $M_{0}$ s.t. $F_{\mathscr{A}}$ is a ($M_{0}$,1,1)-sa function. This means that there exists an M s.t.  $R^{n}$ can be partitioned into M \emph{linear blocks} where $F_{\mathscr{A}}$ is a linear function in each block. Furthermore, $F_{\mathscr{A}}$ must be a Lipschitz function in each block. Since $F_{\mathscr{A}}$ is continuous in $R^{n}$, it is a Lipschitz function in $R^n$, which means there exists an L s.t.
$$|F_{\mathscr{A}}(\vec{x})-F_{\mathscr{y}}| \leq L||\vec{x}-\vec{y}||$$ for any $\vec{x},\vec{y}\in R^{n}$.
Then we can prove Lemma 3.
\par
\textbf{Lemma 3}:\quad If assumption 1 and Condition 1 hold, then for any ray X, if $F_{\mathscr{A}}(\vec{x})$ is constant in X, then
$$F_{\mathscr{A}}(\vec{x})=0$$ for any $\vec{x}$ in X.
\par
Proof of Lemma 3: We assume $F_{\mathscr{A}}$ is L-Lipschitz. For simplicity, let $v=F_{\mathscr{A}}(X)$ and assume $v\geq 0 $ without loss of generality. Then
we define a set $X^{+}=\{\vec{a}:\exists \vec{x}\in X s.t. ||\vec{x}-\vec{a}||\leq \frac{v}{2L}  \}$.
Apparently, $F_{\mathscr{A}}(\vec{x}) \geq v/2$ for any $\vec{x} \in X^{+}$ and the volume of $X^{+}$ is $+\infty$. Thus,
\begin{align}
\int_{R^{n}}|F_{\mathscr{A}}(\vec{x})| &\geq \int_{X^{+}}|F_{\mathscr{A}}(\vec{x})| \\
&\geq \frac{v}{2}\int_{X^{+}}1 \\
&=+\infty
\end{align}

Then we can prove Lemma 2.
\par
Proof of Lemma 2: We prove this lemma with mathematical induction.
\par
\textbf{Basis:} The k=0 case is simple. We let $X_{0}=R^{n}$. It is easy to verify that $S_{i}(0)$ holds for i=1,2,3,4.
\par
\textbf{Inductive step:} Given that $S_{i}(k)$ holds for i=1,2,3,4, we will prove that $S_{i}(k+1)$ holds for i=1,2,3,4 too. Let $X_{k+1}=\{\vec{x}:\vec{x}\in X_{k}\ and \ for\ any \ j=1,2,\dots,n, \ F_{k+1,j,\mathscr{A}}(\vec{x})>0 \}$. Apparently, $X_{k+1}$ is a \emph{linear block} which is a subset of $X_k$. We will prove $X_{k+1}$ satisfies $S_{i}(k+1)$ for i=1,2,3,4.
\par
Based on $S_{4}(k)$, it is easy to see $F_{k+1,\mathscr{A}}$ is a linear function on $X_{k}$.
There exist a $n\times n$ matrix $W_{k+1}$ and a $n\times 1$ vector $b_{k+1}$ such that on $X_{k}$
$$F_{k+1,j,\mathscr{A}}(\vec{x})=W_{k+1}\vec{x}+\vec{b_{k+1}}$$.
We define
$$P_{k+1,i}=\{\vec{x}:W_{k+1}(i,)\vec{x}+\vec{b_{k+1}}(i)>0 \}$$
for $i\in [n]$.
Thus
$$X_{k+1}=\cap_{i=1}^{n}P_{k+1,i}\cap X_{k}$$
Note $P_{k+1,i}$ is convex and $X_{k}$ is convex based on $S_{1}(k)$. Thus $X_{k+1}$ is convex and so that $S_{1}(k+1)$ holds.
\par
Now we are going to prove $S_{2}(k+1)$ holds. For any $\vec{x} \in X_{k}\backslash X_{k+1}$, there exists $j(\vec{x}) \in [n]$, such that
$F_{k+1,j(\vec{x}),\mathscr{A}}(\vec{x})\leq 0$. Note $j(\vec{x})$ depends on $\vec{x}$, but we write it as $j$ for simplicity.

Since $W_{k+1}$ is an $n\times n$ matrix, there must exist an n-dimensional vector
$\vec{\alpha}(\vec{x}) \neq 0$ such that $\vec{\alpha}(\vec{x}) \perp W_{k+1}(i,) \ i\in [n],i\neq j$. Note, $\vec{\alpha}(\vec{x})$ depends on $\vec{x}$, however, we write it as $\vec{\alpha}$ for simplicity. We assume $W_{k+1}(j,)\vec{\alpha}\leq 0 $. If it does not hold, we substitute $-\vec{\alpha}$ for $\vec{\alpha}$.
Then we consider the following set $$IRX_{\vec{x}}=
\{\vec{c}:\vec{c}=\vec{x}+t\vec{\alpha}\in X_{k},t\geq 0 \}$$, the intersection of $X_k$ and the ray corresponding to $\vec{\alpha}$ and $\vec{x}$. By $S_{1}(k)$, $X_k$ is convex. Obviously, the ray corresponding to $\vec{\alpha}$ and $\vec{x}$ is also convex. Thus $IRX_{\vec{x}}$ is a convex set and so that a continuous part of a ray.
For any $\vec{y}\in IRX_{\vec{x}}$ and any $i\in [n],i\neq j$,
\begin{align}
F_{k+1,i,\mathscr{A}}(\vec{y})&=W_{k+1}(i,)(\vec{x}+t\vec{\alpha})+\vec{b_{k+1}}(i) \\
&=W_{k+1}(i,)\vec{x}+\vec{b_{k+1}}(i) \\
&=F_{k+1,i,\mathscr{A}}(\vec{x}),
\end{align}
Thus, for $i\in [n],i\neq j$
\begin{align}
R_{k+1,i,\mathscr{A}}(\vec{y}) &=Relu(F_{k+1,i,\mathscr{A}}(\vec{y})) \\
&=Relu(F_{k+1,i,\mathscr{A}}(\vec{x}))) \\
&=R_{k+1,i,\mathscr{A}}(\vec{x})
\end{align}
Besides, for any $\vec{y}\in IRX_{\vec{x}}$, when i=j,
\begin{align}
F_{k+1,i,\mathscr{A}}(\vec{y})&=W_{k+1}(i,)(\vec{x}+t\vec{\alpha})+\vec{b_{k+1}}(i) \\
&\leq W_{k+1}(i,)\vec{x}+\vec{b_{k+1}}(i) \\
&=F_{k+1,i,\mathscr{A}}(\vec{x})\\
&\leq 0
\end{align}
Thus,when i=j,
\begin{align}
R_{k+1,i,\mathscr{A}}(\vec{y}) &=Relu(F_{k+1,i,\mathscr{A}}(\vec{y})) \\
&=0 \\
&=Relu(F_{k+1,i,\mathscr{A}}(\vec{x}))) \\
&=R_{k+1,i,\mathscr{A}}(\vec{x})
\end{align}
In general, we find
$R_{k+1,\mathscr{A}}$ is constant on $IRX_{\vec{x}}$. Therefore $F_{\mathscr{A}}$ is constant on $IRX_{\vec{x}}$.
We define
$$T=sup\{t:\vec{x}+t\vec{\alpha}\in IRX_{\vec{x}}\}$$
Since $IRX_{\vec{x}}$ is a continuous part of a ray,
$\{t:\vec{x}+t\vec{\alpha}\in IRX_{\vec{x}}\}$ is an interval.
\par
If $T=+\infty$, then $IRX_{\vec{x}}$ is a ray and thus we can conclude $F_{\mathscr{A}}(\vec{x})=0$ by using Lemma 3.
 \par
 If $T<+\infty$, for any $\epsilon >0$, there exist $T_{1},T_{2} $ such that
\begin{align}
&T-\epsilon < T_{1}<T<T_{2}<T+\epsilon \\
&\vec{x}+T_{1}\vec{\alpha} \in X_{k} \\
&\vec{x}+T_{2}\vec{\alpha} \notin X_{k}
\end{align}
By the definition of $B(X_{k})$, $\vec{x}+T\vec{\alpha} \in B(X_{k})$. By $S_{3}(k)$,
$$F_{\mathscr{A}}(\vec{x}+T\vec{\alpha})=0  $$.
On the other hand, $F_{\mathscr{A}}$ is constant on $IRX_{\vec{x}}$. Because of continuity it is constant on $$\overline{IRX_{\vec{x}}}=IRX_{\vec{x}}\cup \{\vec{y}:for \ any \ \epsilon>0, \exists \vec{u}\in IRX_{\vec{x}},||\vec{y}-\vec{u}||<\epsilon  \}$$ Obviously, $\vec{x}+T\vec{\alpha}\in \overline{IRX_{\vec{x}}}$.  Thus,
$$F_{\mathscr{A}}(\vec{x})=F_{\mathscr{A}}(\vec{x}+T\vec{\alpha})$$
Since $F_{\mathscr{A}}(\vec{x}+T\vec{\alpha})=0  $,then
$$F_{\mathscr{A}}(\vec{x})=0$$
In all, for any $\vec{x}\in X_{k}\backslash X_{k+1}$, if assumption 1 holds, $F_{\mathscr{A}}(\vec{x})=0$. Besides, since for any $\vec{x}\in X_{k}^{c}$, $F_{\mathscr{A}}(\vec{x})=0$, then $S_{2}(k+1)$ holds.
\par
Because $F_{\mathscr{A}}$ is continuous and $S_{2}(k+1)$ holds, we can easily find $S_{3}(k+1)$ holds.
\par
By the definition of $X_{k+1}$,
 $$F_{k+1,i,\mathscr{A}}(\vec{x})>0, for \ any \ i\in [n]  \ and \ \vec{x}\in X_{k+1}$$.
 Thus,on $X_{k+1}$,
 \begin{align}
 R_{k+1,i,\mathscr{A}}(\vec{x})&=Relu(F_{k+1,i,\mathscr{A}}(\vec{x}))\\
 &=F_{k+1,i,\mathscr{A}}(\vec{x}) \\
 &=W_{k+1}\vec{x}+\vec{b_{k+1}}
 \end{align}
 It is a linear function. $S_{4}(k+1)$ holds.
 \par
 We finish the proof of Lemma 2.

\par
Proof of Lemma 1: If assumption 1 holds, by setting $k=h-1$ in Lemma 3, we find
there exists a \emph{linear block} $LBX=X_{k}$ such that
\begin{itemize}
\item LBX is convex.
\item $F_{\mathscr{A}}(\vec{x})=0 \ for \ any \ \vec{x}\notin LBX \ or\ \vec{x}
\in B(LBX) $
\item $R_{h-1,\mathscr{A}}$ is a linear function on LBX.
\end{itemize}
Since
$$F_{\mathscr{A}}=A_{h}R_{h-1,\mathscr{A}}+u_{h}$$,
$F_{\mathscr{A}}$ is a linear function on LBX. As $F_{\mathscr{A}}=0$ outside LBX, to finish the proof we just need to prove that
for any $\vec{x}\in LBX$, $F_{\mathscr{A}}(\vec{x})=0$. For any $\vec{x}\in LBX$, let $$L_{\vec{x}}=\{a\vec{x},a\in R\}$$
 and
 $$IL_{\vec{x}}=L_{\vec{x}}\cap LBX$$
Since LBX and $L_{\vec{x}}$ are both convex, $IL_{\vec{x}}$ is convex. Thus there exists an interval A such that
$$t\vec{x}\in IL_{\vec{x}}\Leftrightarrow t\in A$$
Apparently, $F_{\mathscr{A}}(t\vec{x})$ is a linear function of t on A.
Define
$$a=inf \ A$$
$$ b= sup \ A$$.
\par
If $a>-\infty,b<+\infty$,then
$a\vec{x},b\vec{x}\in B(LBX)$. Thus
$$F_{\mathscr{A}}(a\vec{x})=F_{\mathscr{A}}(b\vec{x})=0$$.
Since $F_{\mathscr{A}}(t\vec{x})$ is a linear function, $$F_{\mathscr{A}}(\vec{x})=0$$
\par
If $a>-\infty,b=+\infty$ or $a=-\infty,b<+\infty$, we assume $a=-\infty,b<+\infty$ without loss of generality. Then $F_{\mathscr{A}}(b\vec{x})=0$. If $F_{\mathscr{A}}(\vec{x})\neq 0$, because of the linearity of $F_{\mathscr{A}}$
$$lim_{t\rightarrow -\infty}F_{\mathscr{A}}(t\vec{x})=+\infty \ or \ -\infty$$
Since $F_{\mathscr{A}}(\vec{x})$ is Lipschitz, it contradicts with $\int_{R^{n}}|F_{\mathscr{A}}(\vec{x})|<+\infty$. So $F_{\mathscr{a}}(\vec{x})=0$
\par If $a=-\infty,b=+\infty$, we can prove $F_{\mathscr{A}}(\vec{x})=0$ in a similar way.
\par
In general, $F_{\mathscr{A}}(\vec{x})=0$ for any $\vec{x}\in R^{n}$ if assumption 1 holds.\\

Then obviously Theorem 2 is a direct result of Lemma 1.

\subsection{Proof of Theorem 3}

\begin{proof}
We denote the input by $\vec{x}=(x_1,x_2,...,x_n)$, and the value of the first layer's nodes of $A$ by $y=(y_1,y_2,...,y_m)$, here $m<n$ and let
\[y_i=(b_i+\sum_{j=1}^{m} a_{ij}x_j)^+\]
where $i = 1,2,\cdots,n$, $j=1,2,\cdots,m$.$b_i$ and $a_{ij}$ are parameters of $A$.Since $m<n$, there exists a non-zero vector $x_0$ in $R_0^n$, which satisfies
\[\vec{x}_0 \perp span\{b_1+\sum_{j=1}^{j=m} a_{1j}x_j,\cdots,b_n+\sum_{j=1}^{j=m} a_{nj}x_j\}\]

Since changes along $x_0$ don't affect the first layer of network $A$: $F_{A}$, which is determined by the first layer of $A$ itself, it is constant along $\vec{x}_0$ as a result. Thus $F_{A}$ must be constant along some fixed direction $x_0$. 

Now we can prove that: given f and a fixed unit vector $x_0$, we have a positive $\epsilon$ that for all continuous $F$ which is constant along the direction $x_0$, the $L^1$ distance between $f$ and $F$ is lower bounded by $\epsilon$. Pick two points $a_0$ and $b_0$ along $x_0$ that $f(a_0) < f(b_0)$, due to the continuity of $f$, there exists positive $r$ and $c$ that for all $a$ in $U(a_0, r)$ and $b$ in $U(b_0, r)$, $f(b)-f(a) >c$. Let the lebesgue-measure of $U(a_0, r)$ be $V$, with the triangle inequality $|f(b)-F(b)|+|f(b-b_0+a_0)-F(b-b_0+a_0)| > f(b)-f(b-b_0+a_0) > c$, we can see there exists such an $\epsilon$ which is $>= Vc$.

Then treat $\epsilon$ as a function of $x_0$. Since $\epsilon$ is positive and continuous because $f$ and $F$ are continuous and have compact domain (so any such $F$ is uniformly continuous, then 'rotating' $F$ by a small angle guarantees a small uniform difference, one can easily see $\epsilon$ is continuous now), it has a lower bound over all unit vector $x_0$. Denote this lower bound as $\epsilon^*$, $\epsilon^*$ must be positive because the set of all unit vector $x_0$ is a compact set (see it as the surface of unit ball). Since $F_A$ must be constant along some direction, $\epsilon^*$ is the desired universal constant for all $F_A$.

\end{proof}

\subsection{Proof of Theorem 4}

We first prove the case with input dimension $n=1$, then the extension to  $n>1$ cases is trivial.

\begin{proof}
We will choose $2k^4$ different points $x^{(1)},x^{(2)},\dots,x^{(2k^4)}\in R$ and consider functions represented by ReLU network on them. Here,
$$x^{(i+2k^{2}j)}=2j+1-\frac{2k^{2}-i}{4k^{2}},i=1,2,\dots,2k^{2},j=0,1,\dots,k^{2}-1$$
For any ReLU network $\mathscr{A}$, we define a $2k^{4}$-dimensional vector
$$f_{\mathscr{A}}=(F_{\mathscr{A}}(x^{(1)}),F_{\mathscr{A}}(x^{(2)}),\dots,F_{\mathscr{A}}(x^{2k^{4}}))$$
We will begin our proof by introducing 2 lemmas.

\textbf{Lemma 4:} \quad We define $$E_{0}=\{(a^{(1)},...,a^{(2k^{4})}):0<a^{(i+2k^{2}j)}<\frac{1}{2}a^{(i+1+2k^{2}j)},i=1,2,...,2k^{2}-1,j=0,1,...,k^{2}-1\}$$
$$E_{w}=\{f_{\mathscr{A}}:\mathscr{A}\ is\ a\ ReLU\ network\ with\ width\ 2k^{2},depth\ 3, input\ width\ and\ output\ width\ 1\}$$
Then $$E_{0}\subset E_{w}$$

proof of Lemma 4:\quad

For any $f\in E_{0}$, we will fabric a ReLU network $\mathscr{A}$ with width $2k^{2}$ and depth 3 such that $f=f_{\mathscr{A}}$. Firstly, it is easy to choose appropriate first layer weights and bias to make
$$R_{1,\mathscr{A}}=((x)^{+},(x-1)^{+},\dots,(x-2k^{2}+1)^{+})'$$
Denote the weights and bias of kth layer by $W_{k,\mathscr{A}}$ and $B_{k,\mathscr{A}}$. $W_{k,\mathscr{A}}$ is a matrix and $B_{k,\mathscr{A}}$ is a vector such that
$$F_{k+1,\mathscr{A}}=W_{k,\mathscr{A}}R_{k,\mathscr{A}}+B_{k,\mathscr{A}}$$
Define $F_{2,i,\mathscr{A}}$ to be the function at the $ith$ node in the second layer, which is a piecewise linear function which is linear between any integral points on the x-axis.
It satisfies:
\[F_{2,i,\mathscr{A}}(x^{(i+2k^{2}j)})=a^{(i+2k^{2}j)}\quad
i=1;j=0,1,...,k^{2}-1\]
\[F_{2,i,\mathscr{A}}(x^{(i+2k^{2}j)})=a^{(i+2k^{2}j)}-2a^{(i-1+2k^{2}j)}\quad
i=2;j=0,1,...,k^{2}-1\]
\[F_{2,i,\mathscr{A}}(x^{(i+2k^{2}j)})=a^{(i+2k^{2}j)}-2a^{(i-1+2k^{2}j)}+a^{(i-2+2k^{2}j)} \quad
i=3,4,...,2k^{2};j=0,1,...,k^{2}-1\]
and that
\[F_{2,i,\mathscr{A}}(2j+1-\frac{2k^{2}-i+1}{4k^{2}}) = 0, i=1,2,...,2k^{2};j=0,1,...,k^{2}-1\]
Together with the linearity between integral points on the x-axis, the function represented by the $ith$ node can be uniquely decided. Then we activate those functions by RELU, and add them up to get the final output $f_{\mathscr{A}}$. One can easily check that
\[f_{\mathscr{A}}=(a^{(1)},...,a^{(2k^{4})})\]
Combined with the definition of $E_0$ and $E_w$, we have:
\[E_{0}\subset E_{w}\]

Define
\[\mathscr{F}_k=\{\mathscr{A}:\mathscr{A}\ is\ a\ ReLU\ network\ with\ width\ 2k^{2},depth\ 3, input\ and\ output\ dimension\ 1 ;f_{\mathscr{A}} \in E_0 \}\]

\textbf{Lemma 5:} For any k$\ge$5, only a 0 measure set(Lebesgue measure on the weight and bias space) of the networks in $\mathscr{F}_k$ can be equaled by a deep network whose width $\le k^\frac{3}{2}$ and depth $\le k+2$.

proof of Lemma 5:\quad

We prove a stronger statement: only a 0 measure set(Lebesgue measure on the weight and bias space) of the networks in $\mathscr{F}_k$ can be equaled on specific $2k^4$ different points $x^{(1)},x^{(2)},\dots,x^{(2k^4)}$,by a deep network whose width $\le k^\frac{3}{2}$ and depth $\le k+2$.
Notice the fact that a network with width $d$ and depth $h$ has degree of freedom = $d^2(h-2)+d(h-1)+2d+1$. Define $\mathscr{B}$ to be one of the deep networks, with width $d \le k^{\frac{3}{2}}$ and depth $h \le k+2$.
Let $g_0$ be the function mapping the parameters of the deep network to $f_{\mathscr{B}}$:
$$g_0: R^{d^2(h-2)+d(h-1)+2d+1} \to  R^{2k^4}$$
$$g_0(all \ parameters)=f_{\mathscr{B}}$$.

When $d\le k^{\frac{3}{2}}$ and $h \le k+2$, the degree of freedom of the deep network $\le k^4+k^3 < 2k^4$, and $g_0$ is $C_1$-derivable almost everywhere. Thus, $B$: the set of all $\beta$, which is the solution space of $g_0$ has a zero measure in $R^{2k^4}$ according to Differential Homeomorphism Theorem. In fact, we can implement the original mapping to a new function $g_1$ $$g_1: R^{2k^4} \to R^{2k^4}, g_1(all \ parameters,p_1,...)=g_0(all \ parameters)$$ in the way of adding variables $p_1,p_2,...,p_{2k^4-d^2(h-2)-d(h-1)-2d-1}$ which have no effect on the value of $F$, then the Jacobian of $g_1$ is zero now because the differential of $F$ to $p_i$s is 0, thus by the transform formulation of integration, the measure of the range is zero.
\[m(range(g_1))= \int_{R^{2k^4}}dg_1 =  \int_{R^{2k^4}} \frac{\partial{g_1}}{\partial{\vec{x}}}d\vec{x}=0\]
It's obvious that $m(E_0)>0$, so $E_0\cap range(g_1)$ is a negligible subset in $E_0$ and as a result only a negligible set of the functions in this family of wide networks can be equaled by such deep networks.

Then because all parameters in these deep networks are bounded, we can extend the difference on finite points to integration on input domain.

Apparently, the shape of such a deep network can be denoted by a vector whose $m^{th}$ entry denotes the width of the $m^{th}$ layer except for the output layer. We denote the shape vector of a network $N$ by S(N). Thus for all networks with $h\le k+2$ and $d_m\le k^{1.5}$,
 $$S(N)\in V$$
 $$here \quad V=\{ (w_1,w_2,...,w_h) \vert \quad h\le k+2 \quad and \quad w_m \le k^{1.5} \quad for \quad any \quad m \} $$

\ Denote the all elements of $V$ by $\{ V_j \}$, we only need to prove Lemma 6 as followed,then $n=1$ case is proved directly by setting $\epsilon=min_{j\le |V|}\{\epsilon_{j}\}$:

\textbf{Lemma 6:}\quad For any wide network $N_{w}$  which can't be equaled by deep networks with width $\le k^{1.5}$ and depth $\le k+2$ as above, there exists a $\epsilon_{j} >0$ for all deep network $N_{d}$ with $S(N)=V_{j}$ satisfies
 $$\int_{0}^{2k^{2}} (N_{d}(x)-N_{w}(x))^{2}\geq \epsilon_{j} $$

Set $\epsilon_{j}=inf\{\int_{0}^{2k^{2}} (N_{d}(x)-N_{w}(x))^{2},S(N_{d})=V_{j}\}$
We are going to prove $\epsilon_{j}>0$.
With the conclusion of inequability above and continuity of the function $N_{d} \ and \ N_{w}$, we know for any $$S(N_{d})=V_{j},\int_{0}^{2k^{2}} (N_{d}(x)-N_{w}(x))^{2}>0
$$
Thus, if $\epsilon_{j}=0$ There must be a sequence $N_{d_{i}}$ satisfies $$\int_{0}^{2k^{2}} (N_{d_{i}}(x)-N_{w}(x))^{2}<\frac{1}{i}$$

Since every bounded sequence(here the assumption of parameters' bound is used, so for different choice of $b$, $\epsilon$ changes) has a convergent subsequence and parameters of a network are bounded as well, we can find a subsequence $N_{d_{i_{j}}},j=1,2,\dots$,every parameter of which converges. We define the network they converge to is $\tilde{N}$. Then for any x,
$(N_{d_{i_{j}}}(x)-N_{w}(x))^{2}$ converges to $(\tilde{N}(x)-N_{w}(x))^{2}$. Besides, the values of them are uniformly bounded. Thus, with Dominated Convergence Theorem, we can find
\[\begin{split}
&\int_{0}^{2k^{2}} (\tilde{N}(x)-N_{w}(x))^{2} \\
=& \int_{0}^{2k^{2}} \lim_{j\to \infty} (N_{d_{i_{j}}}(x)-N_{w}(x))^{2} \\
=& {\lim_{j\to \infty}} \int_{0}^{2k^{2}} (N_{d_{i_{j}}}(x)-N_{w}(x))^{2} \\
=& 0 \\
\end{split}\]
This causes contradiction to our conclusion of inequability above. So $\epsilon_{i}>0$ and we are finished with the proof of the case with $n=1$.
\end{proof}

For cases with $n>1$, we denote these $n$ inputs by $x_1,...,x_n$. We construct the same wide network for $x_1$ only and ignore other inputs(set the weights from them to the first later to be 0). Our wide network still has width $2k^2$ and depth 3, and for any deep network with width $\le k^{1.5}$ and depth $\le k+2$ all our results above hold as well (for the choice of the prechosen $2k^4$ points, their value on $x_2,...,x_n$ can be arbitary). The whole proof is finished now.

\end{document}